\documentclass[conference]{IEEEtran}
\IEEEoverridecommandlockouts
\usepackage{cite}
\usepackage{amsmath,amssymb,amsfonts}
\usepackage{graphicx}
\usepackage{textcomp}
\usepackage{xcolor}
\usepackage{hyperref}
\usepackage[utf8]{inputenc}
\usepackage{booktabs} 
\usepackage{siunitx} 
\usepackage{subcaption}
    
\begin{document}

\title{Statistical Analysis of Sentence Structures through ASCII, Lexical Alignment, and PCA\\}

\author{\IEEEauthorblockN{
Abhijeet Sahdev}
\IEEEauthorblockA{
\textit{New Jersey Institute of Technology}\\
as4673@njit.edu}
}

\maketitle

\begin{abstract}
While utilizing syntactic tools such as parts-of-speech (POS) tagging has helped us understand sentence structures and their distribution across diverse corpora, it is quite complex and poses a challenge in natural language processing (NLP). This study focuses on understanding sentence structure balance—usage of nouns, verbs, determiners, etc.—without relying on such tools without relying on such tools. It proposes a novel statistical method that uses American Standard Code for Information Interchange (ASCII) to represent text of 11 text corpora from various sources and their lexical category alignment after using their compressed versions through PCA,  and analyzes the results through histograms and normality tests such as Shapiro-Wilk and Anderson-Darling Tests. By focusing on ASCII codes, this approach simplifies text processing, although not replacing any syntactic tools but complementing them by offering it as a resource-efficient tool for assessing text balance. The story generated by Grok shows near normality indicating balanced sentence structures in LLM outputs, whereas 4 out of the remaining 10 pass the normality tests. Further research could explore potential applications in text quality evaluation and style analysis with syntactic integration for broader tasks.

\end{abstract}


\section{Introduction}
As we progress into the intelligence age \cite{altman2025ia}, celebrating the byproducts of plausible experiments that helped reinforce our faith in deep learning as the guiding methodology for future advancements, addressing  large language models such as ChatGPT \cite{openai2024gpt4technicalreport}, Gemini \cite{geminiteam2024gemini15unlockingmultimodal} , etc., as compressed versions of the Internet that have transformed text generation, yet, one may always encounter grammatical fallacies in their response. There are various approaches to resolve this problem as comprehensively discussed in \cite{omelianchuk2024pillarsgrammaticalerrorcorrection}. However, these methods and other traditional approaches rely on syntactic tools such as parts-of-speech (POS) tagging which again increase the complexity by utilizing word embeddings such as BERT \cite{devlin2019bertpretrainingdeepbidirectional} as we progress further in an NLP pipeline. The process depicted in \cite{churakov2024postagginghighlightskeletalstructure} is also computationally intensive in obtaining the sentence structure.
\\
Our study introduces a novel statistical approach that leverages ASCII codes and Principal Component Analysis which may have been overlooked in recent years to assess sentence structure balance. Earlier approaches of PCA have been limited to classification tasks \cite{inproceedingsJa}. Here, we focus on representing natural language such as English in machine codes such as ASCII. It may not be intuitive to think that texts can be  directly represented as numbers through a standardized mapping such as ASCII. For lexical alignment, seventeen lexical categories that include abbreviation, adjective, adposition, adverb, affix, conjunction, contraction, determiner, interjection, noun, other, phrase, predeterminer, preposition, pronoun, symbol and verb were also converted to ASCII. For example: affix is mapped to [65,102,102,105,120]. 

These lexical categories’ ASCII representations are compressed using PCA. In our setting, the cumulative explained variance was numerically close to 1.0 for the reductions we used, indicating strong redundancy in the representation. Usually, ASCII values are used for basic tasks such as storing text, however, motivated by this observation,  we represented lexical categories as a single vector of 17 dimensions, Each sentence from our text corpora was sent through the same pipeline and finally aligned with this vector and their distributions were analyzed for normality, hypothesizing that balanced text - characterized by harmonious usage of lexical resources exhibited normal distribution. This hypothesis is tested on ten different diverse text sources along with one that was a combination of the previous ten. The sources varied from blogs, news, articles, movie reviews and a story generated by Grok, an LLM from xAI to evaluate its applicability between human and AI-generated texts. The results demonstrate that ASCII can be used for high-level text analysis, expanding its capabilities to statistical linguistic analysis.

\section{Background}
\subsection{\textbf{Principal Component Analysis}}
It aims to project higher dimensional data to lower dimensions while ensuring that the variance in the original data is reflected as much as possible. This is captured using explained variance per component. The following are the steps to compute PCA \cite{shlens2014tutorialprincipalcomponentanalysis} for a given data matrix $Z$. \\
\begin{itemize}
    \item First the data is centered by subtracting the mean of each feature:
\[
Z_{\text{centered}} = Z- \mu
\]
where \( Z \) is the data matrix and \( \mu \) is the mean of each feature (column).

    \item The covariance matrix is calculated as:
\[
\Sigma = \frac{1}{n-1} Z_{\text{centered}}^T Z_{\text{centered}}
\]
where \( n \) is the number of data points.
    \item The covariance matrix \( \Sigma \) is decomposed to find the eigenvectors (principal components) and eigenvalues:
\[
\Sigma a = \lambda a
\]
where \( a \) is an eigenvector, and \( \lambda \) is the corresponding eigenvalue.
\item The eigenvectors are sorted in descending order by their corresponding eigenvalues. The top \( k \) eigenvectors are chosen as the principal components:
\[
A_k = [a_1, a_2, \dots, a_k]
\]
where \( A_k \) is the matrix of the top \( k \) eigenvectors. We set $\bar{\mathbf{K}}$ as one-dimensional or seventeen-dimensional, depending on the use case in our study.

    \item Finally, the data is projected onto the principal components:
    \begin{equation}
        X = Z_{\text{centered}} A_k
        \label{eq:pca}
    \end{equation}
     Where \( X \) is the matrix of the data in the reduced-dimensional space.
\end{itemize}
\subsection{\textbf{Dot Product}}
It helps us measure lexical alignment in our study of arrays with 17 dimensions, it is the sum of the product of individual components, represented as:
\begin{equation}
\bar{\mathbf{J}} \cdot \bar{\mathbf{K}} = \sum_{i=1}^{17} j_i k_i
\label{eq:dot}
\end{equation}

\section{Methodology}
\begin{figure}[htbp]
\includegraphics[width=\columnwidth]{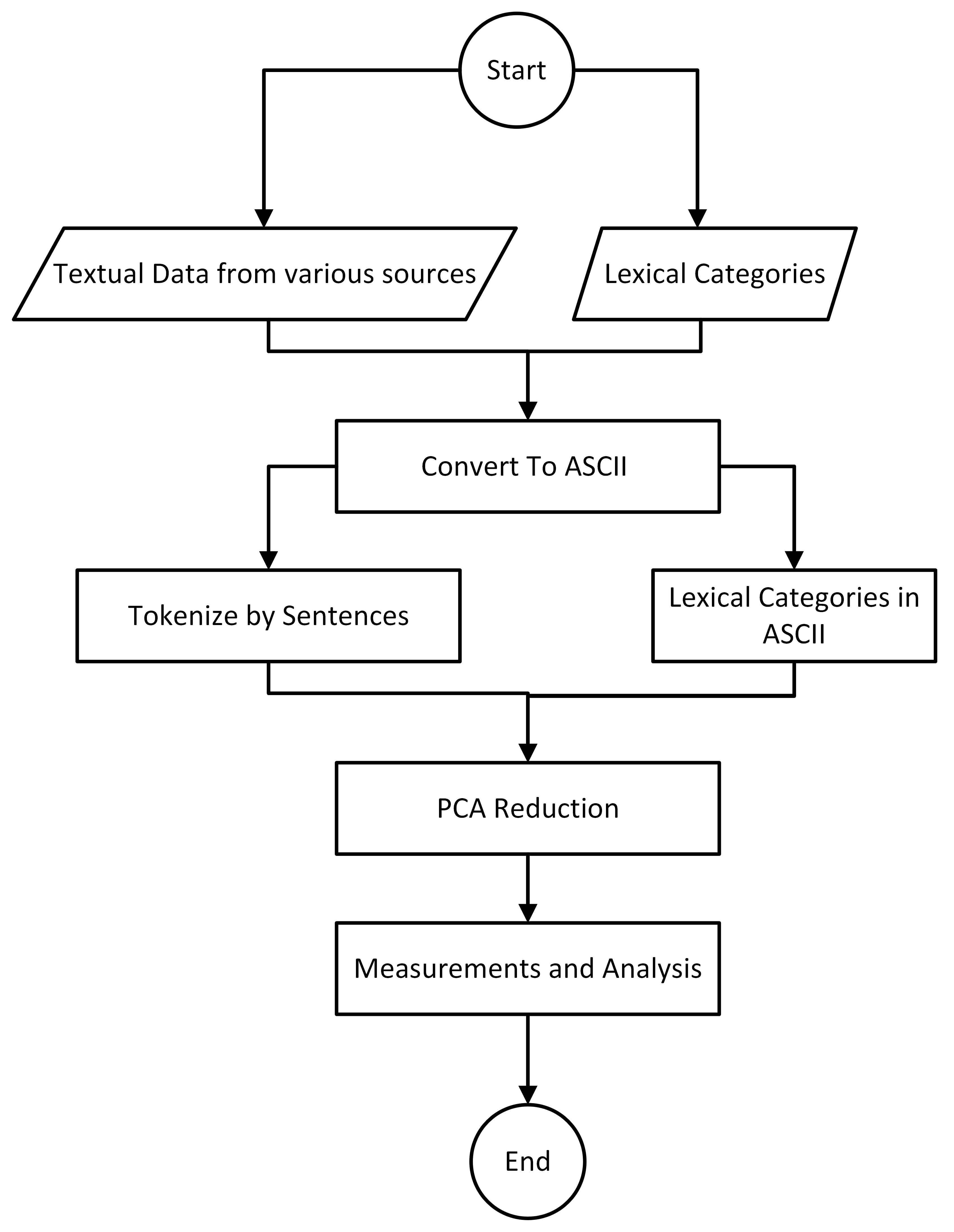}
\caption{Experimental Setup}
\label{fig:1}
\end{figure}

\begin{figure}[htbp]
\includegraphics[width=\columnwidth]{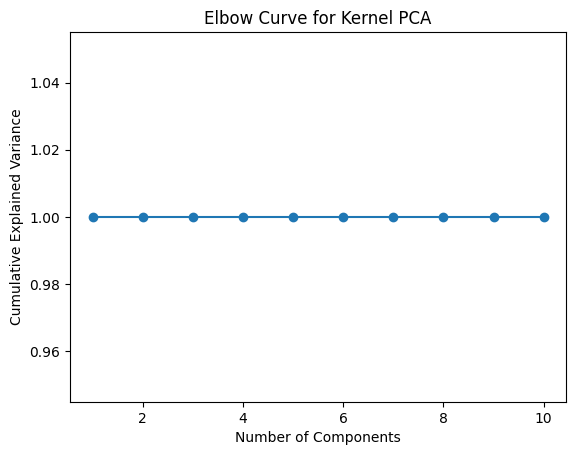}
\caption{Explained Variance}
\label{fig:2}
\end{figure}

\begin{figure}[htbp]
\includegraphics[width=\columnwidth]{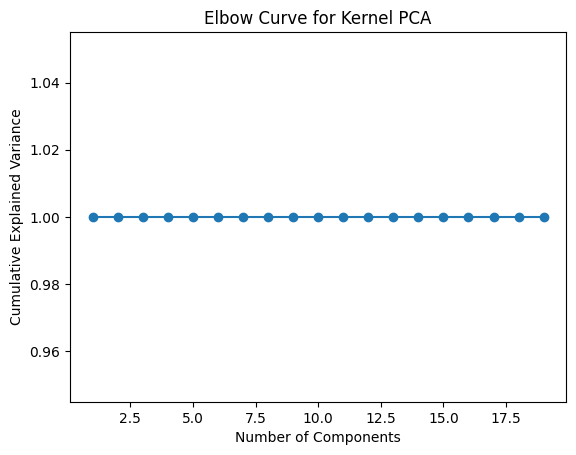}
\caption{Explained Variance for the Entire Text Corpus}
\label{fig:3}
\end{figure}

To compute the above measurements, the experimental setup is designed as shown in \ref{fig:1}. The setup begins by fetching the seventeen lexical categories mentioned earlier and in the same order from an application programming interface. First, lexical categories are converted to ASCII codes and compressed to one dimension using Equation \ref{eq:pca}. For this reduction, the cumulative explained variance was shown to be numerically close to 1.0 (Fig.~\ref{fig:2}), indicating strong redundancy in the ASCII-encoded lexical category representation. Compressed values are recorded here and they act as vector $\bar{\mathbf{K}}$ in Equation \ref{eq:dot}. Next, we selected seven sources: two blogs \cite{altman2025ia,altman2025blog}, three articles \cite{guardian2024,nytimes2025,npr2022}, one story generated using Grok \cite{github2025}, and movie reviews from Hugging Face \cite{Pang+Lee:05a} (train/test/validation). Text corpora from these sources were sent through a pre-processing pipeline to extract sentences in ASCII codes. These ASCII coded sentences were then reduced using Equation \ref{eq:pca} to seventeen dimensions. Across all text corpora, the cumulative explained variance after reduction to 17 dimensions remained numerically close to 1.0, suggesting that the ASCII sentence representations are highly redundant under PCA. 
This behavior is expected given the deterministic and low-entropy nature of ASCII encodings. Explained variance for the entire text corpus (when combined as one, it is a 2-D array of shape (29326,446)) is shown in Figure \ref{fig:3}.  The ASCII corpora, taken one source at a time, tokenized by sentences, converted to ASCII and compressed to seventeen dimensions make up $\bar{\mathbf{J}}$ in Equation \ref{eq:dot}. The dot product was computed and analyzed using measures mentioned in the following sections. The entire jupyter notebook can be viewed at \cite{github2025notebook}.

\section{\textbf{Statistical Measures}}
\subsection{Shapiro-Wilk Test}
We used \cite{eb32428d-e089-3d0c-8541-5f3e8f273532} to assess normality of the dot product distribution, which the Shapiro-Wilk test does by evaluating if a sample is from a normally distributed population by comparing the observed order statistics to expected values under normality. The test statistic \( S \) is as follows:
\begin{equation}
     S = \frac{\left( \sum_{i=1}^{a} n_i g_{(i)} \right)^2}{\sum_{i=1}^{a} (g_i - \bar{g})^2}
     \label{eq:shapiro}
\end{equation}

where \( g_{(i)} \) are the ordered sample values (i.e., the \( i \)-th order statistic), \( g_i \) are the original sample values, \( \bar{g} \) is the sample mean, and \( n_i \) are coefficients derived from the expected values of order statistics of a standard normal distribution. A value of $S$ close to $1$ suggests the data is approximately normal. In practice, we reject normality when the p-value is below $\alpha=0.05$. This test is used on datasets that are small or moderate.

\subsection{Anderson-Darling Test}
The Anderson-Darling Test \cite{dataset} is another normality test that places more emphasis on the tails of the distribution compared to other tests like Shapiro-Wilk. It measures the distance between the empirical cumulative distribution function (\textit{ECDF}) of the sample and the cumulative distribution function (\textit{CDF}) of a normal distribution. The test statistic \( A^2 \) is defined as:
\begin{equation}
    A^2 = -a - \frac{1}{a} \sum_{i=1}^{a} (2i - 1) \left[ \ln(F(t_{(i)})) +  \ln(1 - F(t_{(a+1-i)})) \right]
    \label{eq:adt}
\end{equation}
where \( a \) is the sample size, \( t_{(i)} \) are the ordered sample values, and \( F(t_{(i)}) \) is the CDF of the normal distribution evaluated at \( t_{(i)} \). A larger \( A^2 \) value indicates a greater departure from normality. This test is especially sensitive to deviations in the tails. In practice, as implemented in Python’s \textit{scipy.stats.anderson} function, the test does not directly provide a p-value. Instead, the computed statistic \( A^2 \) is compared to critical values at specified significance levels (e.g., 5\%, corresponding to $\alpha = 0.05$ ). If \( A^2 \) exceeds the critical value, the null hypothesis, i.e, the data follows a normal distribution is rejected, indicating non-normality; otherwise, it is accepted.

\begin{figure}[t]
\centering
\begin{subfigure}[t]{0.48\columnwidth}
    \includegraphics[width=\textwidth]{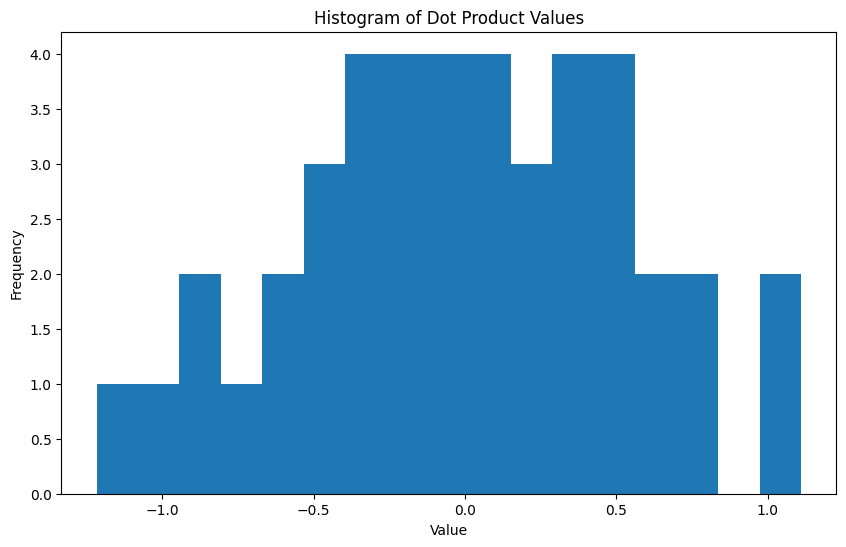}
    \caption{Blog\_ia}
    \label{subfig:image1}
\end{subfigure}
\hfill
\begin{subfigure}[t]{0.48\columnwidth}
    \includegraphics[width=\textwidth]{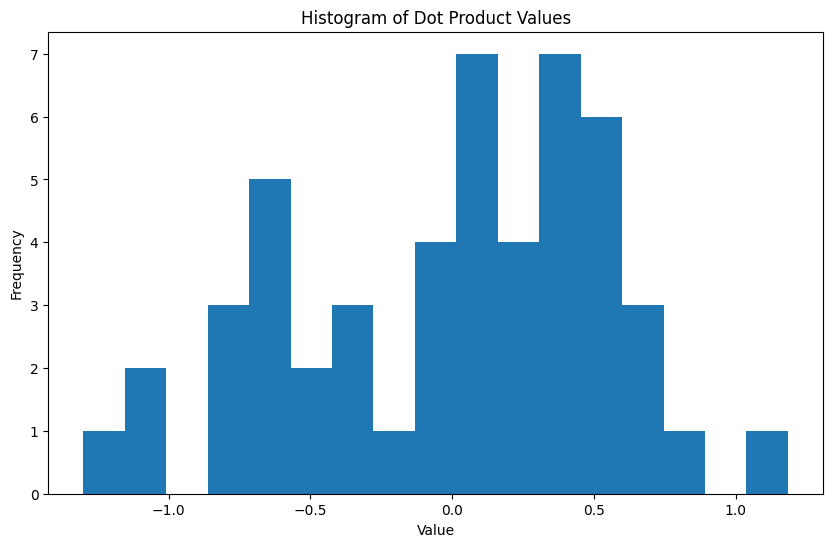}
    \caption{Guardian's article}
    \label{subfig:image2}
\end{subfigure}

\begin{subfigure}[t]{0.48\columnwidth}
    \includegraphics[width=\textwidth]{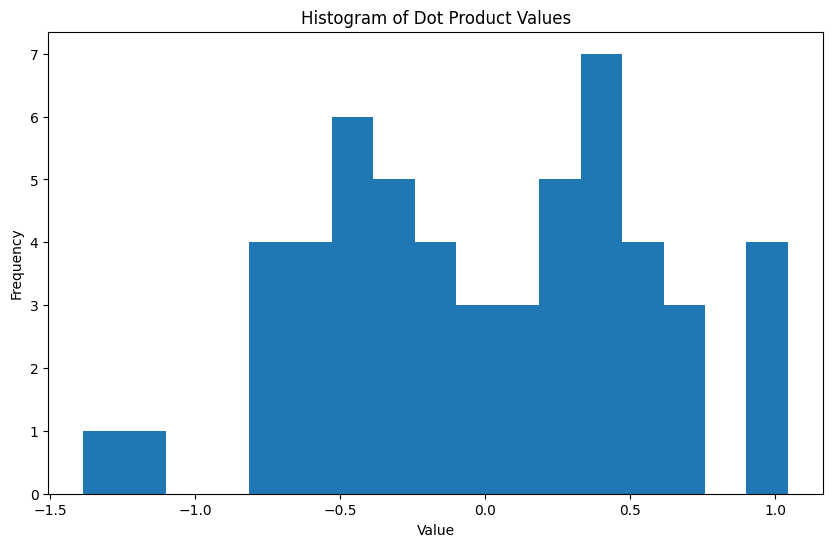}
    \caption{Article}
    \label{subfig:image3}
\end{subfigure}
\hfill
\begin{subfigure}[t]{0.48\columnwidth}
    \includegraphics[width=\textwidth]{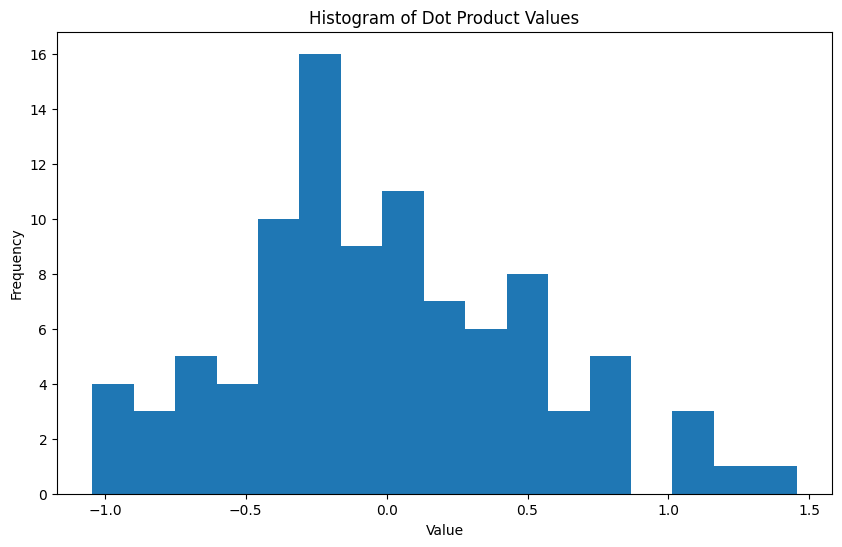}
    \caption{Blog\_ref}
    \label{subfig:image4}
\end{subfigure}

\begin{subfigure}[t]{0.48\columnwidth}
    \includegraphics[width=\textwidth]{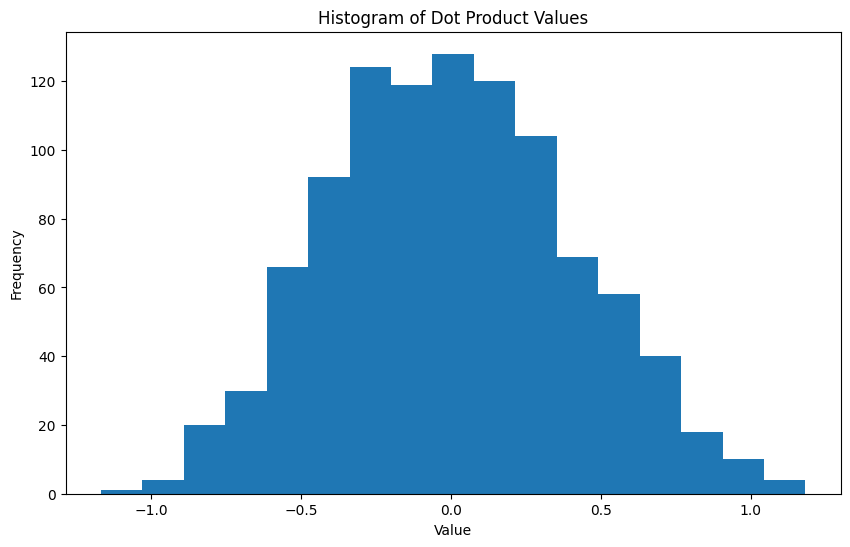}
    \caption{Generated using Grok}
    \label{subfig:image5}
\end{subfigure}
\hfill
\begin{subfigure}[t]{0.48\columnwidth}
    \includegraphics[width=\textwidth]{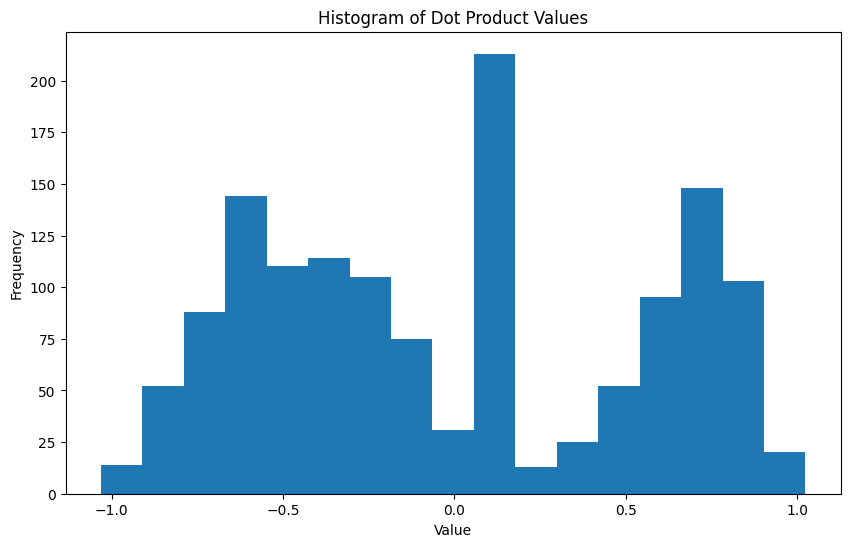}
    \caption{Test Data}
    \label{subfig:image6}
\end{subfigure}

\begin{subfigure}[t]{0.48\columnwidth}
    \includegraphics[width=\textwidth]{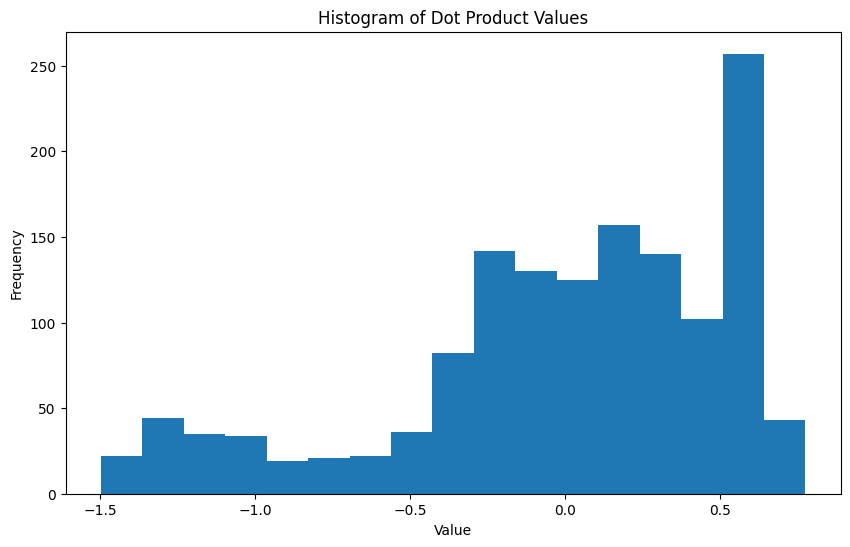}
    \caption{Validation Data}
    \label{subfig:image7}
\end{subfigure}
\hfill
\begin{subfigure}[t]{0.48\columnwidth}
    \includegraphics[width=\textwidth]{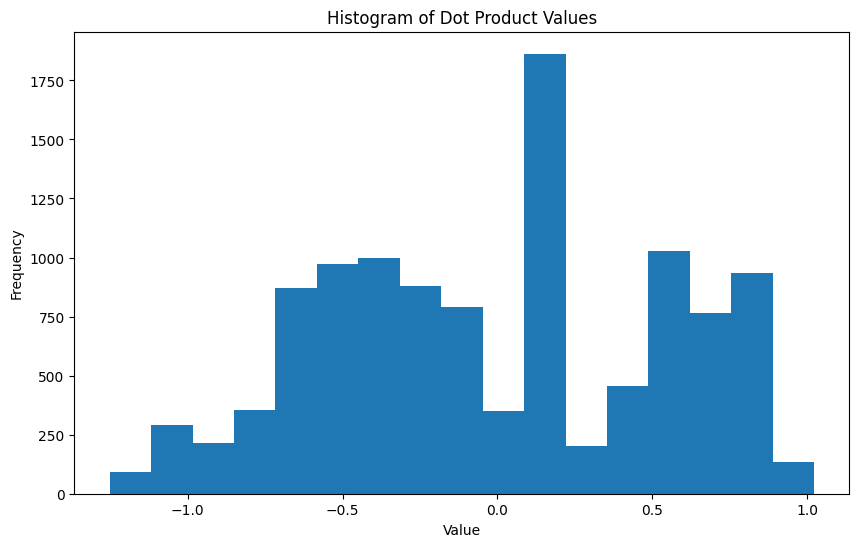}
    \caption{Train Data}
    \label{subfig:image8}
\end{subfigure}

\begin{subfigure}[t]{0.48\columnwidth}
    \includegraphics[width=\textwidth]{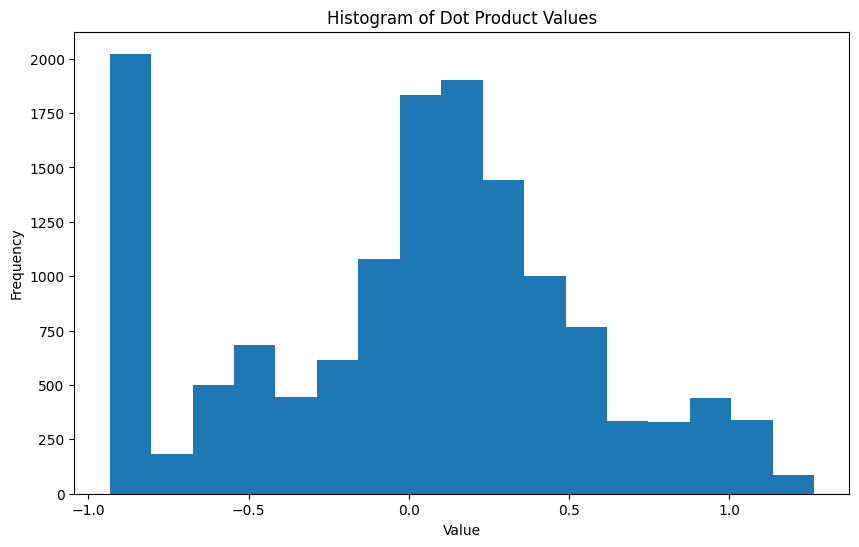}
    \caption{Train, Test and Validation data}
    \label{subfig:image9}
\end{subfigure}
\hfill
\begin{subfigure}[t]{0.48\columnwidth}
    \includegraphics[width=\textwidth]{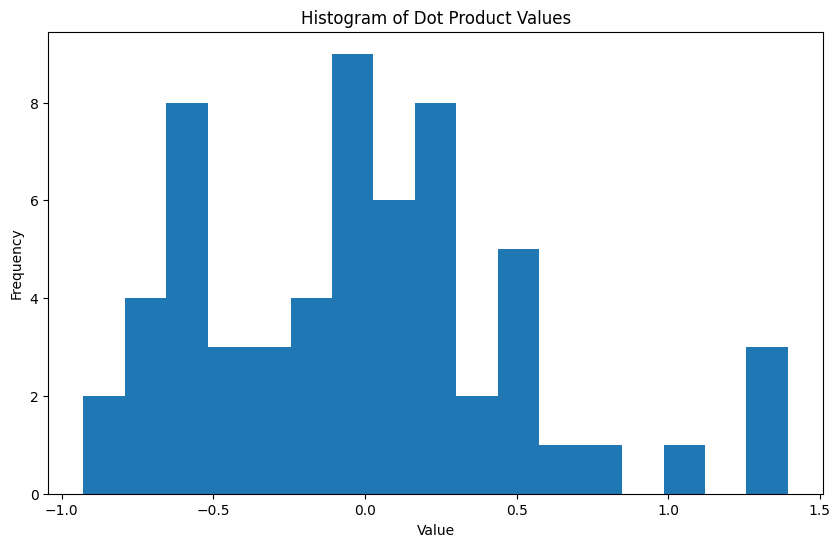}
    \caption{NYT's article}
    \label{subfig:image10}
\end{subfigure}

\caption{Histogram of dot products per experiment}
\label{fig:ten_images_1}
\end{figure}

\begin{table*}[t]
\centering
\caption{Shapiro-Wilk \ref{eq:shapiro} and Anderson-Darling Test \ref{eq:adt} results for Normality Across Experiments}
\label{tab:normality_tests}
\begin{tabular}{l S[table-format=5.0] S[table-format=1.6] S[table-format=1.2e-2] l S[table-format=3.6] S[table-format=1.3] l}
\toprule
Experiment & {Sample Size} & {Shapiro Statistic} & {Shapiro P-Value} & {Shapiro Normal} & {A-D Statistic} & {AD Critical Value (5\%)} & {A-D Normal} \\
\midrule
Blog\_ia & 43 & 0.991321 & 0.983950 & Yes & 0.092526 & 0.729 & Yes \\
Guardian & 50 & 0.964547 & 0.137800 & Yes & 0.757544 & 0.736 & No \\
Mus & 54 & 0.977142 & 0.387746 & Yes & 0.410529 & 0.739 & Yes \\
NYT & 60 & 0.953110 & 0.021862 & No & 0.680246 & 0.743 & Yes \\
Blog\_ref & 96 & 0.979726 & 0.143459 & Yes & 0.604613 & 0.757 & Yes \\
Grok & 1007 & 0.995795 & 0.007514 & No & 1.076552 & 0.784 & No \\
Reviews\_test & 1402 & 0.934546 & 2.47e-24 & No & 30.344235 & 0.785 & No \\
Reviews\_val & 1411 & 0.908294 & 2.53e-28 & No & 35.084494 & 0.785 & No \\
Reviews\_train & 11197 & 0.962414 & 9.56e-47 & No & 130.846467 & 0.787 & No \\
Reviews\_all & 14010 & 0.954351 & 5.00e-54 & No & 191.833786 & 0.787 & No \\
Combined & 29326 & 0.931265 & 1.36e-76 & No & 820.083099 & 0.787 & No \\
\bottomrule
\end{tabular}
\end{table*}

\begin{figure*}[t]
\centering
\begin{subfigure}[t]{0.19\textwidth}
    \includegraphics[width=\textwidth]{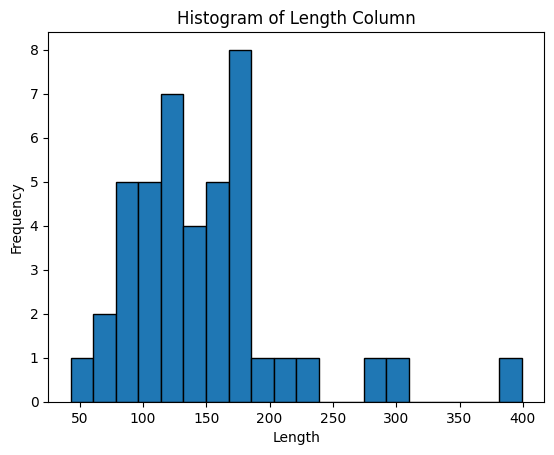}
    \caption{Blog\_ia}
    \label{subfig:image1.1}
\end{subfigure}
\hfill
\begin{subfigure}[t]{0.19\textwidth}
    \includegraphics[width=\textwidth]{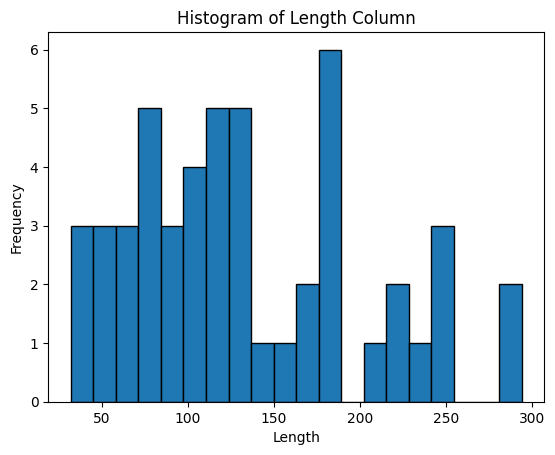}
    \caption{Guardian's article}
    \label{subfig:image1.2}
\end{subfigure}
\hfill
\begin{subfigure}[t]{0.19\textwidth}
    \includegraphics[width=\textwidth]{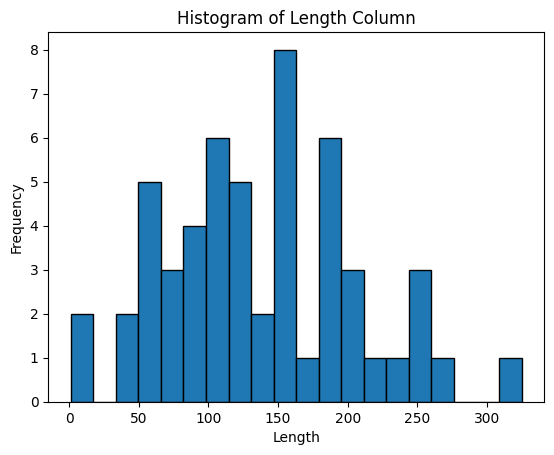}
    \caption{Article}
    \label{subfig:image1.3}
\end{subfigure}
\hfill
\begin{subfigure}[t]{0.19\textwidth}
    \includegraphics[width=\textwidth]{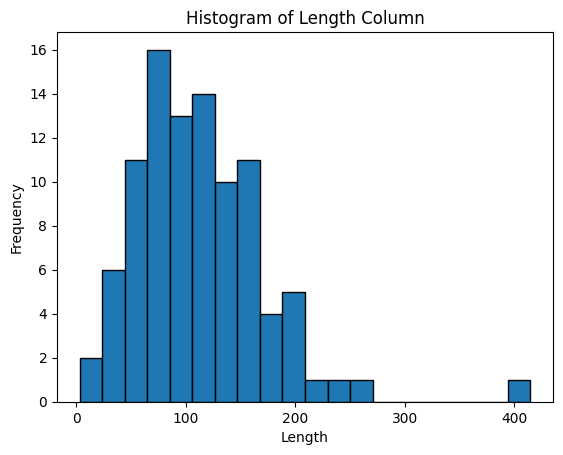}
    \caption{Blog\_ref}
    \label{subfig:image1.4}
\end{subfigure}
\hfill
\begin{subfigure}[t]{0.19\textwidth}
    \includegraphics[width=\textwidth]{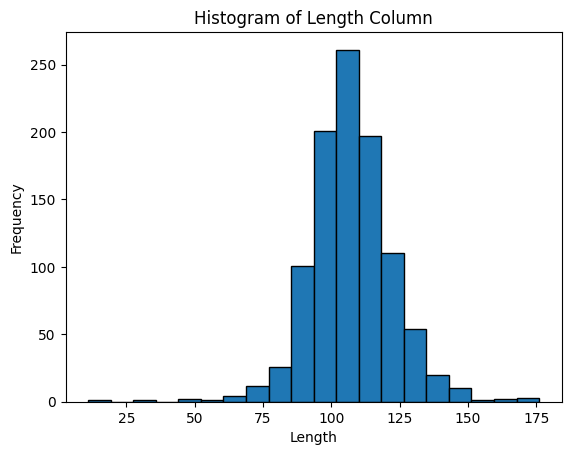}
    \caption{Generated using Grok}
    \label{subfig:image1.5}
\end{subfigure}

\begin{subfigure}[t]{0.19\textwidth}
    \includegraphics[width=\textwidth]{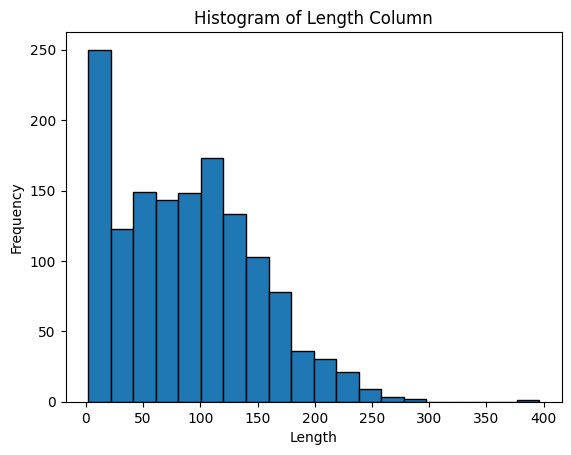}
    \caption{Test Data}
    \label{subfig:image1.6}
\end{subfigure}
\hfill
\begin{subfigure}[t]{0.19\textwidth}
    \includegraphics[width=\textwidth]{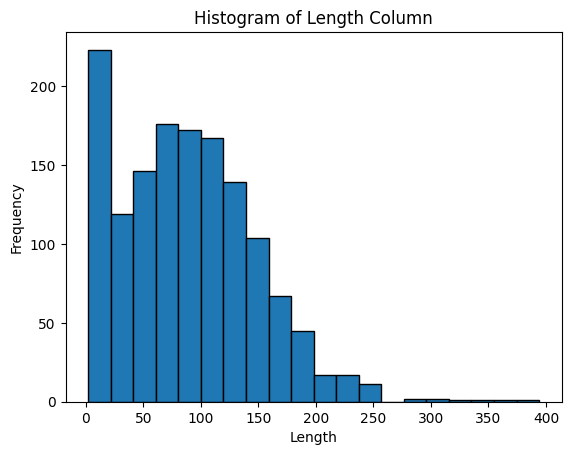}
    \caption{Validation Data}
    \label{subfig:image1.7}
\end{subfigure}
\hfill
\begin{subfigure}[t]{0.19\textwidth}
    \includegraphics[width=\textwidth]{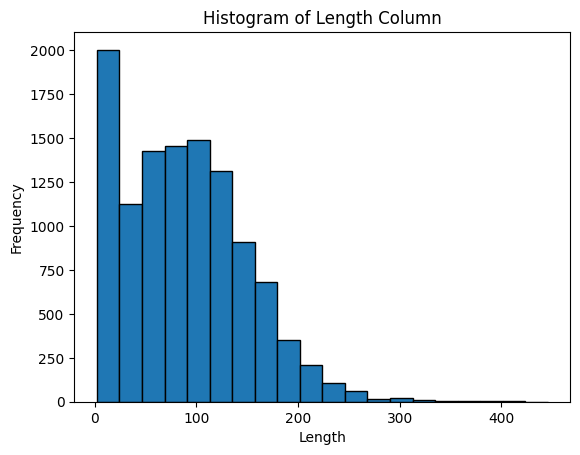}
    \caption{Train Data}
    \label{subfig:image1.8}
\end{subfigure}
\hfill
\begin{subfigure}[t]{0.19\textwidth}
    \includegraphics[width=\textwidth]{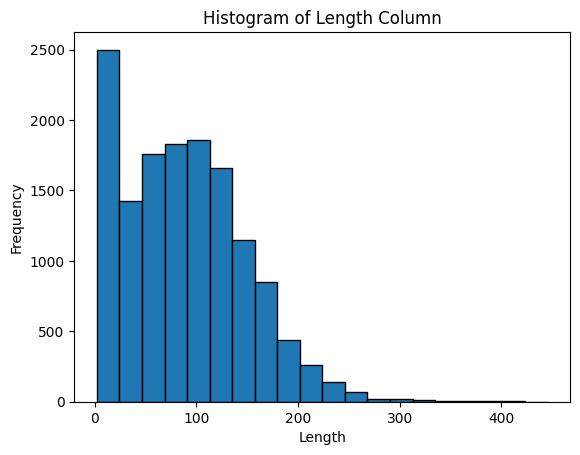}
    \caption{Train, Test and Validation data}
    \label{subfig:image1.9}
\end{subfigure}
\hfill
\begin{subfigure}[t]{0.19\textwidth}
    \includegraphics[width=\textwidth]{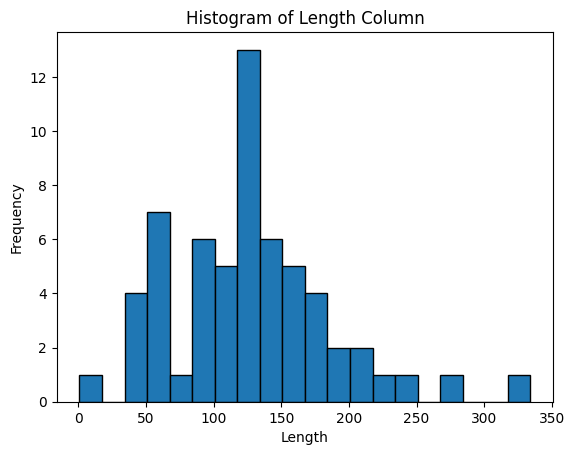}
    \caption{NYT's article}
    \label{subfig:image1.1.10}
\end{subfigure}

\caption{Histogram of length of sentences}
\label{fig:ten_images}
\end{figure*}

\section{Results}
Table \ref{tab:normality_tests} depicts the results of the various experiments conducted along with their tests for normality. In either tests, four experiments clear the normality tests. The LLM generated story fails to clear the normality test after lexical alignment in spite of showing a bell curve in Figure \ref{subfig:image5}. Movie reviews dataset, either train \ref{subfig:image8}, test \ref{subfig:image6}, validation set \ref{subfig:image7}, don't clear the normality tests. Blogs and articles picked up from the Internet clear these normality tests indicating that their structural content in terms of the usage of lexical resources is well balanced.

\section{Analysis and Discussion}
Analyzing the text of \cite{altman2025ia}, we noticed that the variation in sentence structures, i.e, the usage of simple, compound and complex sentences was balanced out throughout the article, which was the motivation behind lexical alignment and can be observed with clearing the normality test from \ref{tab:normality_tests} (Blog\_ia) and Figure \ref{subfig:image1}. Similar trend was observed in \cite{npr2022} which also clears the statistical tests. Notice that the histogram of length of sentences for the text corpus that clear the test have multiple values for it in Fig \ref{fig:ten_images}. Taking Equation \ref{eq:dot} and Figure \ref{fig:ten_images_1} into consideration that highlights that a plethora of samples have a dot product value of zero, indicating orthogonality with the vector $\bar{\mathbf{K}}$, however, we do understand the limitations of such stringent representations. The ones that don't clear the test seem skewed in their distribution of length indicating a limitation in style of content as observed in Figures \ref{subfig:image1.6}, \ref{subfig:image1.7}, \ref{subfig:image8}, \ref{subfig:image9} which could be observed through variations in their dot product graphs as having submodular distributions (see Figures \ref{subfig:image6}, \ref{subfig:image7}, \ref{subfig:image8}, and \ref{subfig:image9}). As for the LLM generated story, viewing Figure \ref{subfig:image1.5} explains that it does not generate a sentence greater than 200 in length indicating no usage of a certain lexical resources too, which is found in Figures \ref{subfig:image1.1}, \ref{subfig:image1.2}, \ref{subfig:image1.3} and \ref{subfig:image1.4}. While the generated content did not have any errors either in spellings or meanings, there are instances where lexical categories occur in unnatural positions such as the usage of preposition 'of' to end a sentence, further research must be done. This approach proposed here complements the existing preprocessing NLP pipeline to analyze text corpora from a different context.

\bibliographystyle{IEEEtran}
\bibliography{references}

\end{document}